%% file: arxiv.tex
\documentclass[10pt,twocolumn,letterpaper]{article}

\usepackage[pagenumbers]{cvpr} 

\input{preamble}
\definecolor{cvprblue}{rgb}{0.21,0.49,0.74}
\usepackage[pagebackref,breaklinks,colorlinks,allcolors=cvprblue]{hyperref}

\def\paperID{4763} 
\def\confName{CVPR}
\def\confYear{2026}

\title{SAVA-X: Ego-to-Exo Imitation Error Detection via Scene-Adaptive View Alignment and Bidirectional Cross View Fusion}

\author{Xiang Li, Heqian Qiu\thanks{Corresponding authors.}, Lanxiao Wang\footnotemark[1], Benliu Qiu, Fanman Meng, Linfeng Xu, Hongliang Li\footnotemark[1]\\
University of Electronic Sience and Technology of China\\
Chengdu, China\\
\scriptsize
\texttt{xianglee@std.uestc.edu.cn,\{hqqiu,lanxiaowang\}@uestc.edu.cn,qbenliu@gmail.com,\{fmmeng,lfxu,hlli\}@uestc.edu.cn}
}

\begin{document}
\maketitle
\begin{abstract}
  Error detection is crucial in industrial training, healthcare, and assembly quality control. Most existing work assumes a single-view setting and cannot handle the practical case where a third-person (exo) demonstration is used to assess a first-person (ego) imitation. We formalize Ego$\rightarrow$Exo Imitation Error Detection: given asynchronous, length-mismatched ego and exo videos, the model must localize procedural steps on the ego timeline and decide whether each is erroneous. This setting introduces cross-view domain shift, temporal misalignment, and heavy redundancy. Under a unified protocol, we adapt strong baselines from dense video captioning and temporal action detection and show that they struggle in this cross-view regime. We then propose \textbf{SAVA-X}, an Align–Fuse–Detect framework with (i) view-conditioned adaptive sampling, (ii) scene-adaptive view embeddings, and (iii) bidirectional cross-attention fusion. On the EgoMe benchmark, SAVA-X consistently improves AUPRC and mean tIoU over all baselines, and ablations confirm the complementary benefits of its components. Code is available at \url{https://github.com/jack1ee/SAVAX}.
\end{abstract}
\section{Introduction}
\begin{figure}[tbp]
  \centering
  \includegraphics[width=1\linewidth]{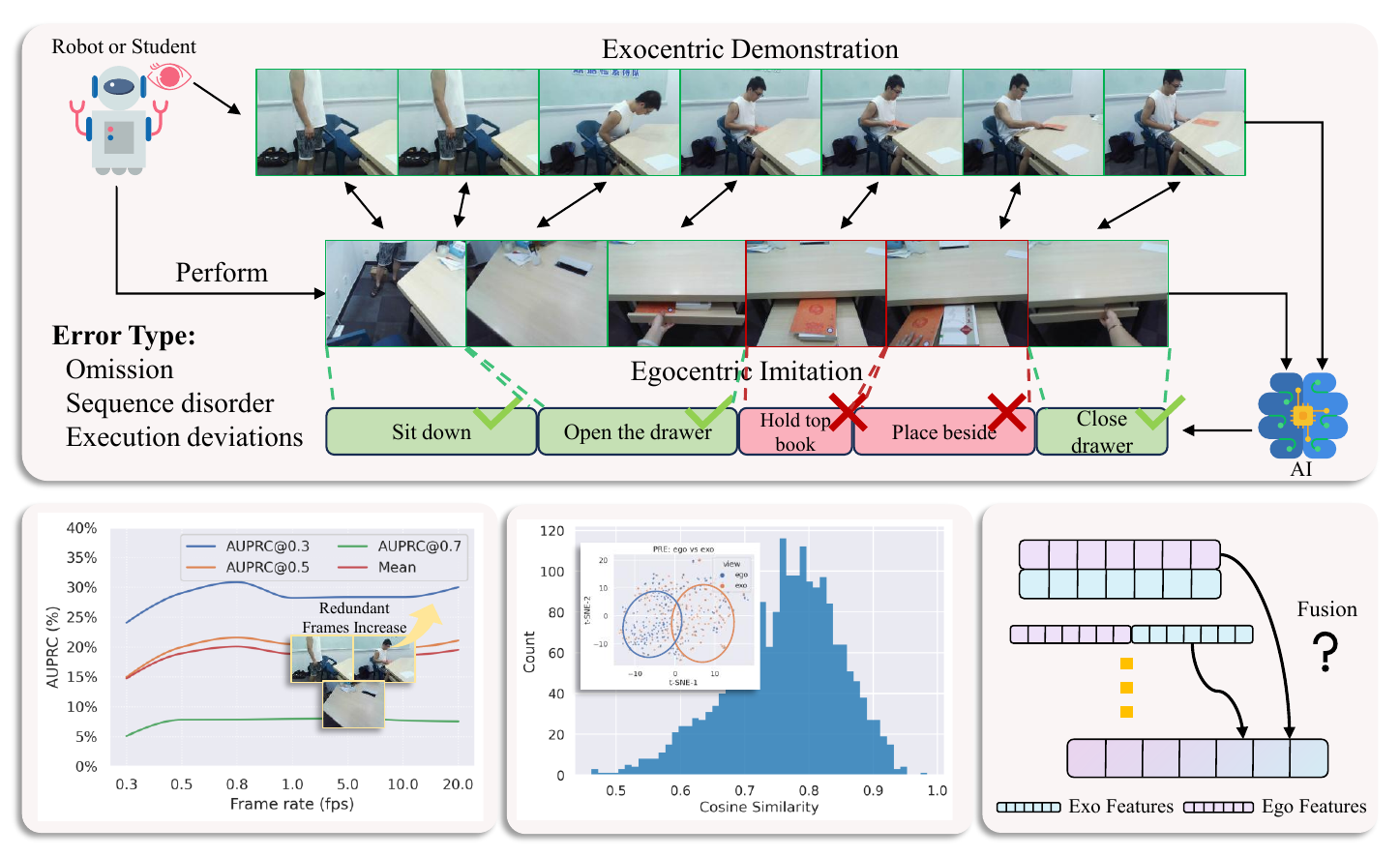} 
  \caption{\textbf{Top}: Schematic of the Ego→Exo imitation-error detection task. The system localizes steps on the ego timeline and judges each step by semantic adherence to the exocentric demonstration, rather than rigid speed/pose matching. \textbf{Bottom-left}: Baseline exhibits a counterintuitive performance drop as the number of input frames increases, partly because redundant frames in the videos introduce distraction. \textbf{Bottom}: There is a pronounced domain shift between Ego and Exo, the distribution of similarities between video-level features of demonstration–imitation pairs is overly dispersed. \textbf{Bottom-right}: A key challenge is how to effectively fuse information from Ego and Exo videos to accomplish the task.}
  \label{fig:intro}
\end{figure}

Wearable cameras and human--robot systems have accelerated research on \emph{egocentric} (first-person) video understanding~\cite{MIR_2025_03_112,plizzari_outlook_2024,fan_benchmarks_2024}. Egocentric videos capture fine-grained hand--object interactions and operator intent, supporting applications such as procedural training, skill assessment, and execution monitoring. While recent progress spans egocentric action classification~\cite{kundu_probres_2025,zhang_masked_2024,shiota_egocentric_2024,schoonbeek_industreal_2024,flaborea_prego_2024,annakukleva_xmic_2024,wang_learning_2023,plizzari_what_2023,gong_mmgego4d_2023,yan_multiview_2022,plizzari_e2_2022,min_integrating_2021} and temporal detection~\cite{yuhanshen_progressaware_2024,reza_hat_2024,quattrocchi_synchronization_2024,liu_endtoend_2024,shi_tridet_2023,zhang_actionformer_2022,wang_egoonly_2023,huang_improving_2020}, most existing approaches assume \emph{egocentric-only} inputs.

In many real-world settings---industrial assembly following an instructor, a nurse imitating a medical protocol, or a robot learning from demonstration---the reference is a \emph{third-person (exocentric)} demonstration, and the goal is to determine whether a first-person execution faithfully imitates it. This cross-view formulation is rarely explored. Prior error-detection studies~\cite{mazzamuto_gazing_2025,flaborea_prego_2024} operate within a single view, leaving a fundamental question unanswered: \emph{how can we detect procedural mistakes when demonstration and execution belong to different, unaligned viewpoints?}

We formalize this problem as \textbf{Ego$\rightarrow$Exo imitation error detection}. Given an exocentric demonstration $V^{exo}$ and an egocentric execution $V^{ego}$ recorded asynchronously and with potentially different durations, the system must (i) localize procedural steps on the ego timeline and (ii) classify each step as correct or erroneous relative to the demonstration. We build upon EgoMe~\cite{qiu_egome_2025}, the only dataset providing paired but unaligned ego--exo videos with fine-grained procedural and error annotations.

This cross-view setting poses three tightly coupled challenges. \textbf{Temporal misalignment.} Ego/exo videos differ in timing, pace, and execution style; duration mismatch is not itself an error, yet it disrupts na{\"i}ve feature alignment. \textbf{Heavy redundancy.} Long videos contain substantial non-informative content~\cite{wang_seal_2025,buch_flexible_2025}, diluting attention mechanisms~\cite{vaswani_attention_2017} and amplifying false positives. \textbf{Pronounced view-domain gap.} Egocentric views emphasize local hand--object interactions, whereas exocentric views capture global posture and scene layout. Their appearance and motion statistics differ significantly~\cite{luo_viewpoint_2025,xue_learning_2023}, causing direct feature fusion to be unreliable.

Addressing these intertwined issues requires more than simply concatenating ego/exo features or adapting single-view temporal detectors. To this end, we propose \textbf{SAVA-X} (Scene-Adaptive View Alignment with Bidirectional Cross View Fusion), a unified \emph{Align--Fuse--Detect} framework that (i) selectively retains informative segments to stabilize alignment under timing mismatch, (ii) conditions representations on learnable view-aware embeddings to reduce domain discrepancy, and (iii) performs complementary cross-view interaction to unify global demonstration cues with local egocentric evidence. Importantly, SAVA-X is designed so that each component targets one of the core challenges while jointly reinforcing the others.

Under a unified evaluation protocol, we adapt strong baselines from dense video captioning and temporal action detection and show that they struggle in this cross-view regime. On EgoMe\cite{qiu_egome_2025}, SAVA-X achieves consistent improvements in AUPRC and mean $\mathrm{tIoU}$ across multiple thresholds. Extensive ablations verify that each design element contributes meaningfully and that their combination yields robust cross-view error detection.

\textbf{Contributions.} (1) We introduce and formalize the new task of \textbf{Ego$\rightarrow$Exo imitation error detection}, highlighting practical significance and core challenges. (2) We propose \textbf{SAVA-X}, an \emph{Align--Fuse--Detect} framework designed to jointly handle temporal misalignment, redundancy, and cross-view domain shift. (3) We establish a unified protocol, adapt strong baselines, and show that SAVA-X delivers consistent gains and complementary component effects on the EgoMe\cite{qiu_egome_2025} benchmark.

\begin{figure*}[t]
  \centering
  \includegraphics[width=0.95\linewidth]{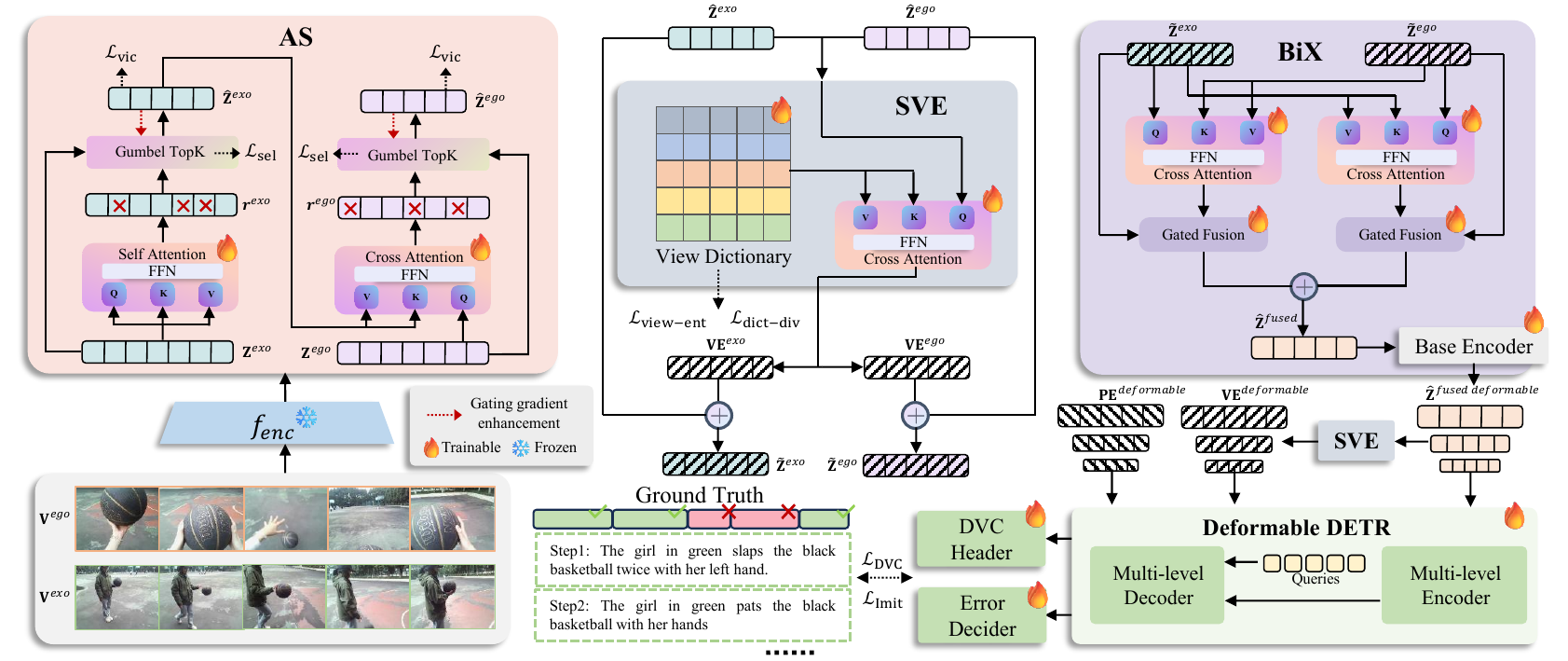} 
  \caption{Overview of SAVA-X. (1) A frozen video encoder extracts per-frame features from the exocentric demonstration and egocentric imitation streams. We apply gated adaptive sampling (Sec. \ref{sec:sampling})—hard Top-K with residual gating, using self-attention scoring for Exo and Exo-conditioned cross-attention scoring for Ego to select key segments. (2) We inject scene-aware dictionary view embeddings (Sec. \ref{sec:view-embed}) together with temporal positions (multi-level), regularized by attention-entropy and prototype-diversity terms, to mitigate cross-view domain shifts. (3) We perform bidirectional cross-attention fusion (Sec. \ref{sec:bixattn}) with learnable gating to align and aggregate complementary cues, yielding a fused sequence that a deformable Transformer encoder–decoder converts into first-person temporal spans and imitation-correctness predictions. Training uses Hungarian set prediction with \(\mathcal{L}_{\text{DVC}}\) and \(\mathcal{L}_{\text{Imit}}\).}
  \label{fig:overview}
\end{figure*}

\section{Related Work}
\textbf{Video Understanding}. Modern video foundation models have evolved from 3D CNNs to large-scale Transformers pretrained with self-supervision. A common downstream paradigm is frozen backbone and lightweight task head. Recent methods design temporal structures or supervision strategies to learn general spatiotemporal representations \cite{timesformer_2021, mvit_2021, videoswin_2022, videomaev2_2023}. For egocentric pretraining, \cite{lin_egocentric_2022, pramanick_egovlpv2_2023, fan_textguided_2024} extend multimodal and self-supervised approaches to first-person scenarios. To handle redundancy in long videos, adaptive frame selection and sparse distillation improve efficiency under fixed token budgets \cite{tang_adaptive_2025, buch_flexible_2025, zou_languageaware_2024}. We similarly employ learned sampling for redundancy and cross-view temporal alignment. With frozen backbones, prompt- or embedding-based adaptation \cite{jia2022visual, ren2025vpt} provides efficient specialization. For multi-view conditioning, \cite{liu2022petr, li2025cameras} integrate geometric priors into Transformers. Inspired by this, we introduce a scene-aware dictionary of view embeddings to encode ego/exo differences and enhance cross-scene generalization.
\\\textbf{Temporal Action Localization}. TAL aims to detect action boundaries and assign class labels in untrimmed videos. Recent work has moved from proposal-based pipelines to single-stage Transformer or query-based formulations, achieving strong benchmark performance \cite{zhang_actionformer_2022, shi_tridet_2023, liu_endtoend_2024}. In egocentric settings with frequent view changes, training directly on egocentric data alleviates domain transfer issues \cite{wang_egoonly_2023}, while unsupervised alignment using synchronized ego–exo pairs further bridges the gap \cite{quattrocchi_synchronization_2024}. For procedural online scenarios, error-free prototypes support error detection, progress-aware segmentation enhances streaming recognition, and memory-augmented designs improve efficiency \cite{shih-polee_error_2024, yuhanshen_progressaware_2024, reza_hat_2024}.
\\\textbf{Dense Video Captioning}. DVC aims to localize multiple events and generate textual descriptions for untrimmed videos \cite{qasim2025dense}. Since its introduction by \cite{krishna2017dense}, research has advanced from coarse activity captioning to fine-grained procedural domains such as cooking, supported by datasets like YouCook2, YouMakeup, and large narrated corpora \cite{nakamura2021sensor, zhou2018towards, wang2019youmakeup, miech2019howto100m}. The dominant paradigm has shifted from two-stage proposal–caption pipelines \cite{xu2019joint, zhou2018end} to end-to-end formulations with set prediction and parallel decoding \cite{wang_endtoend_2021, yang2023vid2seq, zhou2024streaming}, reducing heuristic dependencies. For cross-view egocentric captioning, \cite{ohkawa_exo2egodvc_2024} establish an exo→ego transfer benchmark using view-invariant adversarial learning to address viewpoint mismatch.

\section{Method}
We propose a unified \emph{SAVA-X} framework (see Fig.~\ref{fig:overview}) that tackles three core challenges in \emph{ego–exo} imitation error detection via complementary designs at the sampling, representation, and fusion levels. We first formalize the task in Sec.~\ref{sec:problem}. In sections~\ref{sec:sampling}, \ref{sec:view-embed} and \ref{sec:bixattn}, we respectively describe the three modules we propose.

\subsection{Problem Formulation}\label{sec:problem}
The Ego–Exo imitation-error detection task is defined as follows. Given a third-person demonstration video and a first-person imitation video that are recorded independently (with unaligned timelines and possibly different durations), the model aims to \emph{localize} procedural step segments on the first-person timeline and \emph{decide} whether each localized segment is a correct imitation. Let
\(
V^{exo}=\{I^{exo}_t\}_{t=1}^{T_x},
V^{ego}=\{I^{ego}_t\}_{t=1}^{T_y}
\)
denote the exocentric and egocentric videos with lengths \(T_x\) and \(T_y\), respectively. The task produces a set prediction
\(
\widehat{\mathcal{D}}=\big\{(\hat t^{st}_n,\hat t^{ed}_n,\hat y_n)\big\}_{n=1}^{N}
\)
, where \(\hat t^{st}_n,\hat t^{ed}_n\in[0,1]\) are the normalized start and end times on the first-person timeline and \(\hat y_n\in\{0,1\}\) denotes the imitation correctness label.

To leverage large pretrained video backbones while reducing training cost \cite{timesformer_2021, mvit_2021, videoswin_2022, videomaev2_2023, alwassel_tsp_2021}, we adopt a frozen pretrained encoder (shared or separate) to extract per-frame or per-segment features. Denote the feature dimension by \(d\) and the frozen encoder by \(f_{\mathrm{enc}}\):
\begin{equation}
  \begin{aligned}
  \mathbf{Z}^{exo}=f_{\mathrm{enc}}(V^{exo})\in\mathbb{R}^{T_x\times d}\\
  \mathbf{Z}^{ego}=f_{\mathrm{enc}}(V^{ego})\in\mathbb{R}^{T_y\times d}
  \end{aligned}
\end{equation}
To suppress redundancy and increase the density of informative segments, we compute saliency scores on each stream and select Top-\(K\) segments to get the resampled sequences \(\hat{\mathbf{Z}}^{exo}\) and \(\hat{\mathbf{Z}}^{ego}\).
Notably, the scorer for the demonstration (exo) uses only \(\mathbf{Z}^{exo}\) to identify key demonstration frames, whereas the scorer for the imitation (ego) conditions on both \(\mathbf{Z}^{ego}\) and the resampled demonstration features \(\hat{\mathbf{Z}}^{exo}\) to obtain cues for imitation correctness and to facilitate better temporal alignment. The resulting sparse sequence lengths \(K_x\) and \(K_y\) satisfy \(K_x<T_x\) and \(K_y<T_y\).

We then augment the sparse sequences with temporal position embeddings and view-condition embeddings to provide the model with relative timing and viewpoint information:
\begin{equation}
  \begin{aligned}
  \tilde{\mathbf{Z}}^{exo}=\hat{\mathbf{Z}}^{exo}+\mathbf{PE}^{exo}+\mathbf{VE}^{exo}\\
  \tilde{\mathbf{Z}}^{ego}=\hat{\mathbf{Z}}^{ego}+\mathbf{PE}^{ego}+\mathbf{VE}^{ego}
  \end{aligned}
\end{equation}
where \(\mathbf{PE}\) denotes vectorial absolute temporal position embeddings and \(\mathbf{VE}\) denotes scene-adaptive view embeddings generated from a shared dictionary. The enhanced sequences \(\tilde{\mathbf{Z}}^{exo}\) and \(\tilde{\mathbf{Z}}^{ego}\) are fused into an Ego–Exo representation \(\tilde{\mathbf{Z}}^{fused}\) used to localize action segments on the first-person timeline and to predict imitation correctness.

For efficient global modeling and cross-sequence interaction over long inputs, the fused representation is processed by a deformable transformer encoder–decoder\cite{zhu_deformable_2021,wang_endtoend_2021}. The decoder employs \(N\) learnable queries to perform iterative refinement. Each decoder query yields one first-person candidate prediction. During training we use a set-matching loss to establish a one-to-one correspondence between predicted and ground-truth segments, and jointly optimize the DVC loss \(\mathcal{L}_{DVC}\) \cite{wang_endtoend_2021} and the imitation-error classification loss \(\mathcal{L}_{Imit}\), detailed in supplementary material.

\subsection{Adaptive Sampling}\label{sec:sampling}
In cross-view imitation scenarios, both the \textit{Exo} and \textit{Ego} streams contain substantial redundancy that is irrelevant for discrimination. However, naively sparsifying them risks discarding details crucial for action localization and error judgment, and further ignores the temporal reference required for cross-view alignment. To address this, we propose \emph{gated adaptive sampling}: during training we generate \textbf{hard indices} via a Gumbel Top$K$ \cite{jang2017categorical} straight-through estimator, while simultaneously applying a \textbf{residual gating} to the full-length features so as to provide the scorer with an \emph{additional differentiable path}, thereby strengthening gradient signals and stabilizing the learning of selection. This strategy preserves discrete selection to ensure downstream modules process only a small set of key moments, yet avoids the gradient sparsity problem of purely hard sampling.

For \(\mathbf{Z}^{exo}\), scores are produced by self-attention followed by a FFN head. Then scores are performed by GumbelTopK \(\mathcal{G}_{k}\) to get hard indices \(\boldsymbol{l}_x\) and soft indices \(\boldsymbol{s}_x\).
\begin{equation}
  \begin{gathered}
\boldsymbol{r}^{exo}=\mathrm{FFN}\!\Big(\mathrm{SelfAttn}(\mathbf{Z}^{exo})\Big)\in\mathbb{R}^{T_x}\\
\big(\,\boldsymbol{l}_x,\,\boldsymbol{s}_x\,\big)=\mathcal{G}_{k}(\boldsymbol{r}^{exo})\\
\end{gathered}
\end{equation}


To further strengthen gradients, we perform residual gating \cite{Jeong_2025_CVPR}
\begin{equation}
  \begin{gathered}
\mathbf{g}^{exo}=\mathbf{1}+\alpha\big(\,\mathrm{Norm}(\boldsymbol{s}_x)-\mathbf{1}\,\big)\\
\hat{\mathbf{Z}}^{exo}=\mathrm{Gather}\big(\,\mathbf{g}^{exo}\odot \mathbf{Z}^{exo},\,\boldsymbol{l}_x\,\big)
  \end{gathered}
\end{equation}
where \(\alpha\in(0,1]\) controls the gating strength, and “Norm” rescales soft indices to have \(\mathrm{mean}\approx 1\). \textbf{Crucially}, the \emph{sequence} fed to downstream modules comes from the \emph{hard indices}. This design makes the loss depend on the soft indices \(\boldsymbol{s}_x\) explicitly, thus providing the scorer with stable gradients.

Ego-side scoring should be sensitive to demonstrative key points. We therefore use the Exo summary \(\hat{\mathbf{Z}}^{exo}\) as keys/values and compute Ego cross-attention scores:
\begin{equation}
  \begin{gathered}
\boldsymbol{r}^{ego}=\mathrm{FFN}\!\Big(\mathrm{CrossAttn}\big(\mathbf{Z}^{ego},\,\hat{\mathbf{Z}}^{exo}\big)\Big)\in\mathbb{R}^{T_y}
  \end{gathered}
\end{equation}
and then apply the same pipeline to produce \(\hat{\mathbf{Z}}^{ego}\).

To prevent selection collapse and representational redundancy, we add a \emph{selection-entropy} regularizer \(\mathcal{L}_{\mathrm{sel}}\)~\cite{pereyra2017regularizing} to the selection distributions, encouraging coverage rather than concentrating probability mass on a few positions, and attach VICReg-style~\cite{bardes2022vicreg} \emph{variance lower bound} and \emph{off-diagonal covariance} penalties \(\mathcal{L}_{\mathrm{vic}}\) to the gated active tokens to suppress dimensional collinearity and collapse. 
\(
\mathcal{L}_{\mathrm{sel}}\) and \(\mathcal{L}_{\mathrm{vic}}
\) are detailed in supplementary material.
In practice, this “hard selection + residual gating” combination improves focus on critical segments and stabilizes cross-view alignment and error detection downstream.

\subsection{Scene-aware Dictionary View Embeddings}\label{sec:view-embed}
Cross-view (Ego/Exo) videos exhibit systematic differences in appearance, composition, and motion statistics \cite{luo_viewpoint_2025, ohkawa_exo2egodvc_2024, xue_learning_2023, quattrocchi_synchronization_2024}: ego-centric footage typically focuses on hand–object interactions and sees less of the global scene, whereas exo-centric footage provides full-body and scene structure. If one directly aligns and fuses frozen features, the model can mistake view-domain shifts for action differences, degrading localization and error detection. Injecting the view condition as learnable tokens \cite{jia2022visual, ren2025vpt}—analogous to positional embeddings—is an effective remedy. However, fixed view embeddings may fail to adapt across diverse scenarios. To this end, we explicitly model the view condition and inject it into feature or attention computation in a scene-adaptive manner, thereby mitigating domain shifts and promoting cross-view alignment and evidence aggregation.

We maintain a shared view–scene dictionary:
\[
\mathbf{D}\in\mathbb{R}^{M\times d},
\]
whose rows capture common view-related sub-factors (e.g., “close hand–object interaction” and “full-body motion structure”). For a stream \(u\in\{\text{ego},\text{exo}\}\) with per-frame features \(\hat{\mathbf{Z}}^{u}\in\mathbb{R}^{T\times d}\) used as queries, we interact with the dictionary via multi-head attention under temperature \(\tau\) to obtain an adaptive view embedding:
\begin{equation}
\mathbf{VE}^{u}=\mathrm{CrossAttn}\!\Big(\tfrac{\hat{\mathbf{Z}}^{u}}{\tau},\ \mathbf{D}\Big)
\end{equation}
To influence the representation deeply without introducing substantial parameters, we inject at two sites:
(i) Pre-fusion injection: inject the view condition once into each Ego/Exo stream before fusion and at the encoder input, so that within-domain alignment is performed first.
(ii) Multi-layer injection in the encoder: at each temporal level of the base encoder’s output, perform another view-embedding injection to realize multi-level modulation.

To ensure the view embedding \emph{meaningfully affects attention allocation} without becoming overly peaky, we regularize toward the uniform distribution—equivalently, we maximize normalized entropy \cite{pereyra2017regularizing}:
\begin{equation}
  \mathcal{L}_{\text{view-ent}}=\frac{1}{\log M}\,\mathbb{E}_{t}\!\Big[\mathrm{KL}\big(\alpha_t\ |\ U_M\big)\Big]
\label{eq:view_ent_kl}
\end{equation}
where \(\alpha_t\in\mathbb{R}^M\) is the attention distribution of the \(t\)th time position to the dictionary, and \(U_M\) is a uniform distribution in M dimensions.
To broaden the dictionary’s coverage and suppress redundancy among prototypes, we first apply \(\ell_2\) normalization to each row \(\mathbf{d}_m\) of \(\mathbf{D}\) to obtain \(\widehat{\mathbf{D}}\). We then minimize the deviation from the identity to encourage approximate orthogonality:
\begin{equation}
\mathcal{L}_{\text{dict-div}}
= \big\|\,\widehat{\mathbf{D}}\,\widehat{\mathbf{D}}^{\top}-\mathbf{I}_M\,\big\|_F^2
\label{eq:dict_div}
\end{equation}
Compared with using only fixed token-type biases, the attention-based dictionary can adaptively emphasize the appropriate view subspace under different scenes. Combined with multi-layer injection, it consistently narrows the Ego/Exo domain gap and provides clearer, more transferable representations for subsequent bidirectional cross-fusion and temporal alignment.

\subsection{Bidirectional Cross-Attention Fusion}\label{sec:bixattn}
After obtaining the two sparse sequences augmented with temporal positions and view conditions, \(\tilde{\mathbf{Z}}^{exo}\in\mathbb{R}^{K_x\times d}\) and \(\tilde{\mathbf{Z}}^{ego}\in\mathbb{R}^{K_y\times d}\), we seek robust semantic alignment and complementary evidence aggregation between unaligned, different-length Ego/Exo streams. One-way conditioning can introduce bias: using only Exo to guide Ego under-covers hand–object details in the first-person view, and the converse holds as well. Therefore, we adopt symmetric bidirectional cross-attention \cite{lu2019vilbert, tan2019lxmert, lee2023cast} so that the two streams retrieve from and constrain each other at the feature level, while residual mixing preserves native per-view representations—balancing “alignment capacity” with “view-specific robustness”.

We compute in parallel:
\begin{equation}
\begin{gathered}
\mathbf{E}^{\star}=\mathrm{CrossAttn}\big(\tilde{\mathbf{Z}}^{ego},\,\tilde{\mathbf{Z}}^{exo}\big)\\
\mathbf{X}^{\star}=\mathrm{CrossAttn}\big(\tilde{\mathbf{Z}}^{exo},\,\tilde{\mathbf{Z}}^{ego}\big)
\end{gathered}
\end{equation}
where \(\mathbf{E}^{\star}\) denotes globally structured/boundary evidence retrieved from the demonstration (Exo), and \(\mathbf{X}^{\star}\) denotes hand–object/detail/context evidence retrieved from the imitation (Ego). 

To prevent either side from overwhelming the other, we apply learnable, gated \cite{Jeong_2025_CVPR} residual mixing that retains only the necessary cross-view gains:
\begin{equation}
  \begin{gathered}
\mathbf{F}^{ego}=(1-\boldsymbol{\gamma}^{e})\,\tilde{\mathbf{Z}}^{ego}+\boldsymbol{\gamma}^{e}\,\mathbf{E}^{\star}\\
\mathbf{F}^{exo}=(1-\boldsymbol{\gamma}^{x})\,\tilde{\mathbf{Z}}^{exo}+\boldsymbol{\gamma}^{x}\,\mathbf{X}^{\star}
\end{gathered}
\end{equation}
with \(\boldsymbol{\gamma}^{e}=\sigma\!\big(\mathbf{W_e}[\tilde{\mathbf{Z}}^{ego};\mathbf{E}^{\star}]\big)\) and \(\boldsymbol{\gamma}^{x}=\sigma\!\big(\mathbf{W_x}[\tilde{\mathbf{Z}}^{exo};\mathbf{X}^{\star}]\big)\), where \(\sigma\) is the sigmoid and \([\cdot;\cdot]\) denotes concatenation. This mixing encourages the model to rely more on cross-view evidence near action boundaries and key interactions, while preserving view-stable representations in background/redundant regions.

We finally add the two gated features to fused features for subsequent decoding:
\begin{equation}
  \tilde{\mathbf{Z}}^{fused}=\tfrac{1}{2}\big(\mathbf{F}^{ego}+\mathbf{F}^{exo}\big)
\end{equation}
Compared with one-way fusion, bidirectional cross-attention imposes complementary semantic constraints: Exo\(\rightarrow\)Ego strengthens first-person boundary cues and step ordering, while Ego\(\rightarrow\)Exo contributes object/hand details and local causality \cite{xue_learning_2023, li_egoexo_2021, quattrocchi_synchronization_2024}. After gating, these are aggregated into \(\tilde{\mathbf{Z}}^{fused}\), which retains per-view stability while gaining cross-view corroboration, thereby facilitating the subsequent query-style decoding to more easily localize erroneous segments and determine their types.
\begin{table*}[t]\small
  \centering
  \begin{tabular*}{0.95\textwidth}{@{\extracolsep{\fill}} l*{5}{c}
    @{}@{\hspace{8pt}}!{\vrule width .8pt}@{\hspace{2pt}}@{}
    *{5}{c}}
  \toprule
  \multirow{2}{*}[-0.8ex]{\textbf{Method}}
   & \multicolumn{5}{c}{\textbf{AUPRC on Validation}}
   & \multicolumn{5}{c}{\textbf{AUPRC on Test}} \\
  \cmidrule(lr){2-6} \cmidrule(lr){7-11}
   & 0.3 & 0.5 & 0.7 & Mean & tIoU
   & 0.3 & 0.5 & 0.7 & Mean & tIoU \\
  \midrule
  \multicolumn{11}{@{\extracolsep{\fill}}l}{\emph{Dense Video Captioning (DVC) baselines}} \\
  PDVC~\cite{wang_endtoend_2021}
   & 28.21 & 20.48 & 7.95 & 18.88 & 58.58
   & 25.74 & 18.08 & 4.79 & 16.20 & 57.98 \\
  Exo2EgoDVC~\cite{ohkawa_exo2egodvc_2024}
   & 31.33 & 20.27 & 7.49 & 19.69 & 59.06
   & 26.26 & 16.30 & 5.42 & 15.99 & 58.15 \\
  \midrule
  \multicolumn{11}{@{\extracolsep{\fill}}l}{\emph{Temporal Action Localization (TAL) baselines}} \\
  ActionFormer~\cite{zhang_actionformer_2022}
   & 31.37 & 15.41 & 2.63 & 16.47 & 48.89
   & 26.96 & 12.88 & 2.40 & 14.08 & 48.25 \\
  TriDet~\cite{shi_tridet_2023}
   & 30.04 & 14.61 & 2.44 & 15.70 & 49.05
   & 26.27 & 13.16 & 1.89 & 13.77 & 49.02 \\
  \midrule
  \multicolumn{11}{@{\extracolsep{\fill}}l}{\emph{Only Egocentric Input}} \\
  PDVC~\cite{wang_endtoend_2021}
   & 19.35 & 13.91 & 5.11 & 12.79 & 57.63
   & 21.40 & 15.10 & 5.33 & 13.94 & 57.19 \\
  \midrule
  \multicolumn{11}{@{\extracolsep{\fill}}l}{\emph{Ours}} \\
  SAVA-X
   & \textbf{33.56} & \textbf{24.04} & \textbf{9.48} & \textbf{22.36} & \textbf{59.31}
   & \textbf{29.37} & \textbf{19.86} & \textbf{6.26} & \textbf{18.50} & \textbf{58.32} \\
  \bottomrule
  \end{tabular*}
  \caption{Comparison on EgoMe validation and test split. Left: results on \emph{validation set}. Right: results on the \emph{test set}. We report AUPRC for the error class at multiple tIoU thresholds (0.3, 0.5, 0.7), their mean, and standalone temporal IoU (tIoU) for localization quality.}
  \label{tab:egome_all_vs_erroronly}
\end{table*}
\section{Experiments}
\subsection{Experimental Settings}
\textbf{Dataset.} To match our Ego→Exo imitation error detection setting, all training and evaluation are conducted on the EgoMe \cite{qiu_egome_2025} dataset. To our knowledge, EgoMe is the only large-scale dataset that simultaneously offers asynchronously captured egocentric and exocentric views in imitation scenarios and provides annotations of erroneous imitation steps. It contains 7,902 pairs of asynchronous Exo–Ego videos (approximately 82.8 hours). In our experiments, we use only RGB videos together with fine-grained procedural step annotations and error labels. We follow the official train/validation/test split with 4,777/997/2,128 video pairs and report all main results and ablation studies on this partition.
\\\textbf{Evaluation Metrics.} Ego$\rightarrow$Exo imitation error detection is essentially a temporal object detection task: the system must both localize segment boundaries and determine whether a segment constitutes an error. Because error segments are typically sparse and class-imbalanced, and our primary concern is detecting errors, we adopt the area under the precision--recall curve (\textbf{AUPRC}) for the error class as the primary metric, which reduces threshold sensitivity and is more informative under imbalance. We report AUPRC for error segments evaluated at multiple temporal Intersection-over-Union (tIoU) thresholds, $\{0.3, 0.5, 0.7\}$, along with their mean. In addition, we report average \textbf{tIoU} \cite{fujita2020soda} separately to isolate localization quality.
\subsection{Implementation Details}
We use TSP \cite{alwassel_tsp_2021} pretrained on ActivityNet \cite{caba2015activitynet} as a frozen feature extractor $f_{\mathrm{enc}}$, to aggregate temporal context from video and obtain $\mathbf{Z}^{\mathrm{exo}}$ and $\mathbf{Z}^{\mathrm{ego}}$, with a unified feature dimensionality of $d=512$. TSP is pretrained with temporal sensitivity for localization-oriented objectives (e.g., temporal action localization and captioning), making it a suitable, task-agnostic foundation for downstream localization/description. In \textbf{SAVA-X}, the hidden dimensionality of all three submodules—adaptive sampling, scene-adaptive view embeddings, and bidirectional cross-attention fusion—is set to $512$. Self-attention and cross-attention each use a single layer, and the feed-forward networks (FFNs) employ a hidden size of $2048$. Following bidirectional fusion, the deformable transformer encoder–decoder stack, the dense video captioning (DVC) head implementation, and the relative weighting of the constituent losses within $\mathcal{L}_{\mathrm{DVC}}$ strictly follow the public PDVC configuration \cite{wang_endtoend_2021}. The imitation discrimination loss weight is set to $\lambda_{\mathrm{Imit}}=0.5$. Other regularization terms (e.g., selection entropy, decorrelation) are weighted within $[0.01,\,0.05]$. We optimize with AdamW \cite{loshchilovdecoupled}, using a batch size of $16$ and a learning rate of $1.0\times 10^{-4}$.
\\\textbf{Baselines.} To establish a robust and reproducible comparative baseline for Ego$\rightarrow$Exo imitation error detection, we select representative methods from two related lines of work and retrain them on EgoMe under a unified setup: from dense video captioning (DVC), \textbf{PDVC} \cite{wang_endtoend_2021} and \textbf{Exo2EgoDVC} \cite{ohkawa_exo2egodvc_2024}; and from temporal action localization (TAL), \textbf{ActionFormer} \cite{zhang_actionformer_2022} and \textbf{TriDet} \cite{shi_tridet_2023}. Given that EgoMe provides asynchronously captured yet coarsely time-aligned ego/exo videos, we adopt a uniform \emph{simple fusion} strategy across all baselines: we concatenate frozen Ego and Exo features along the channel dimension and feed the resulting representation into each method. Then additional error detection head is inserted. Except for these changes, all other architectural components and hyperparameters strictly follow the original configurations of the respective methods.

\subsection{Results}
\begin{figure}[htbp]
  \centering
  \includegraphics[width=1\linewidth]{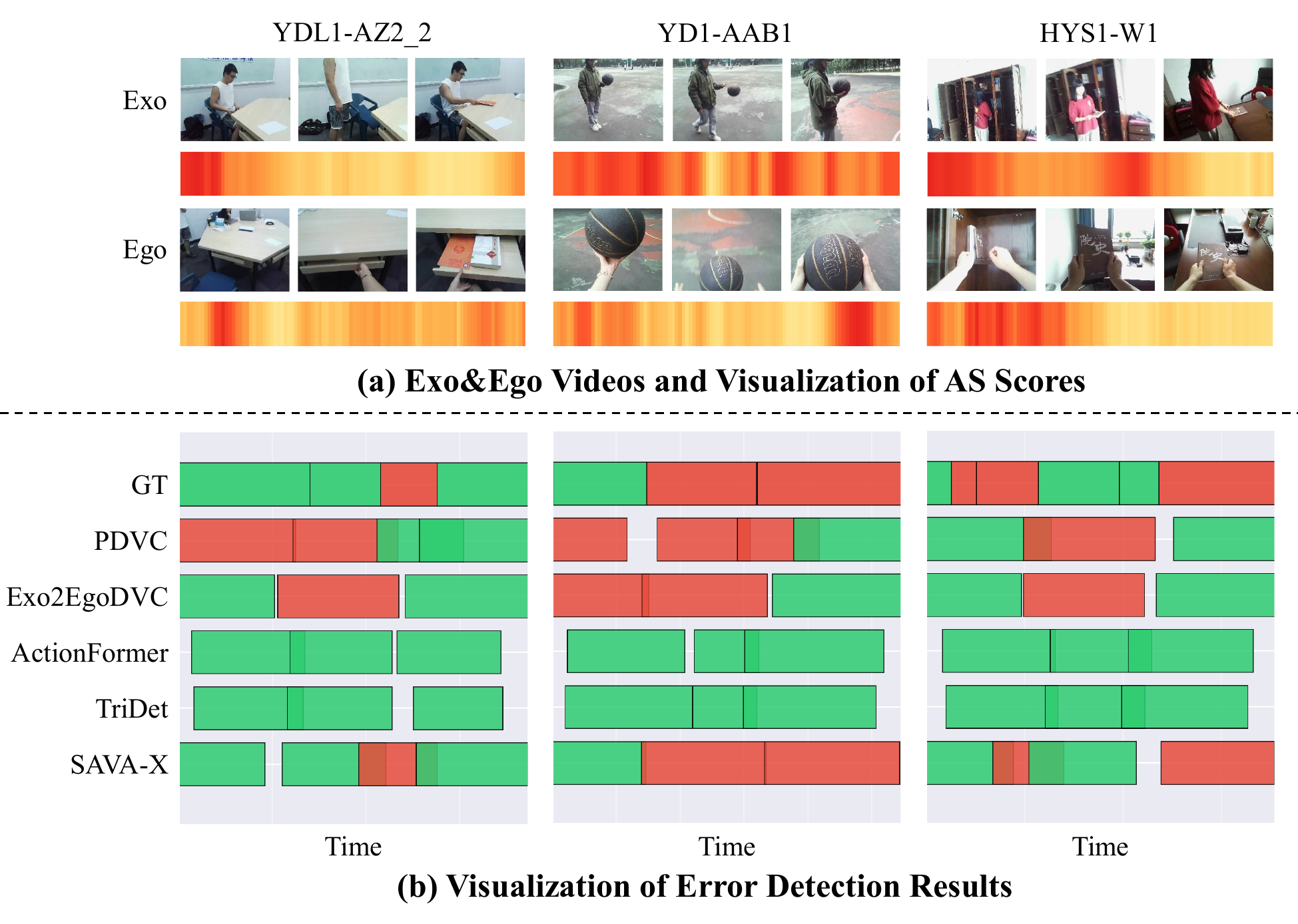} 
  \caption{Qualitative visualization examples of Ego to Exo imitation error localization. (a): Exocentric demonstration and egocentric imitation with corresponding frame saliency maps. The deeper the red, the more significant. (b): Ground truth (GT) and baseline vs. SAVA-X. Red represents error steps while green represents right steps.}
  \label{fig:performance}
\end{figure}

\textbf{Quantitative Comparison on EgoMe.} Table~\ref{tab:egome_all_vs_erroronly} reports performance on the EgoMe validation and test sets for multiple baselines and our SAVA-X framework. SAVA-X attains the best AUPRC and tIoU across all thresholds. On the validation set, SAVA-X achieves a Mean AUPRC of 22.36, an absolute \gain{+2.67} (relative \gain{+13.56\%}) improvement over the strongest baseline Exo2EgoDVC (19.69). Localization quality (tIoU) also shows a modest increase, underscoring the effectiveness and potential of our approach. The test set exhibits consistent trends. Fig.~\ref{fig:performance}b visualizes the results of our models. Overall, SAVA-X delivers simultaneous gains under both stringent high-threshold regimes (hard detections) and coverage-oriented low-threshold regimes, indicating a strong capacity to capture fine-grained cues of erroneous actions. We also report PDVC under single-view input, which degrades performance markedly, indicating that third-person demonstrations are crucial for localizing procedural steps and reducing false positives, thereby validating the task design.
\\\textbf{Ablation Study.} Table~\ref{tab:ablation_savx_nolabel} presents a comprehensive ablation demonstrating the effectiveness of each component. (i) \emph{Single module.} All three modules yield consistent gains: AS, SVE, and BiX improve by \gain{+10.70\%}/\gain{+12.76\%}/\gain{+11.55\%} over the unmodified backbone. These results confirm that redundancy removal with salient-segment amplification (AS), domain-gap mitigation (SVE), and cross-view bi-directional evidence fusion (BiX) each independently enhance error-step identification. (ii) \emph{Pairwise combinations.} SVE+BiX achieves the highest performance, clearly surpassing other pairs, highlighting the impact of narrowing the domain gap and mutual cross-checking; AS+SVE is strongest at medium/high thresholds, suggesting that de-redundancy and view adaptation sharpen boundary precision; AS+BiX brings a more moderate gain, indicating susceptibility to domain shift and noise prior to explicit view conditioning. (iii) \emph{All three combined.} SAVA-X attains the best overall performance, demonstrating strong complementarity among the three modules.
\begin{table}[!htbp]\small
  \centering
  \setlength{\tabcolsep}{4pt}
  \begin{tabular}{cccccccc}
  \toprule
  \multicolumn{3}{c}{\textbf{Modules}} & \multicolumn{4}{c}{\textbf{AUPRC}} & \multirow{2}{*}[-0.8ex]{\textbf{tIoU}} \\
  \cmidrule(lr){1-3}\cmidrule(lr){4-7}
  \textbf{AS} & \textbf{SVE} & \textbf{BiX} & 0.3 & 0.5 & 0.7 & Mean &  \\
  \midrule
   &  &  & 28.21 & 20.48 & 7.95 & 18.88 & 58.58 \\
  \cmark &  &  & 30.90 & 22.60 & 9.21 & 20.90 & 58.88 \\
   & \cmark &  & 31.64 & 22.87 & 9.37 & 21.29 & 59.27 \\
   &  & \cmark & 33.08 & 21.86 & 8.23 & 21.06 & 58.27 \\
  \cmark & \cmark &  & 30.89 & \textbf{24.26} & \textbf{10.32} & 21.82 & 58.96 \\
  \cmark &  & \cmark & 29.98 & 22.27 & 8.70 & 20.32 & 58.14 \\
   & \cmark & \cmark & \textbf{35.09} & 22.58 & 9.31 & 22.33 & 58.76 \\
  \cmark & \cmark & \cmark & 33.56 & 24.04 & 9.48 & \textbf{22.36} & \textbf{59.31} \\
  \bottomrule
  \end{tabular}
  \caption{Ablation on EgoMe validation split without a variant label column. AS: Adaptive Sampling; SVE: Scene-Adaptive View Embedding; BiX: Bidirectional Cross-Attention Fusion.}
  \label{tab:ablation_savx_nolabel}
\end{table}

\subsection{Component Analysis}
\subsubsection{AS Analysis}
\begin{figure}[htbp]
  \centering
  \includegraphics[width=0.9\linewidth]{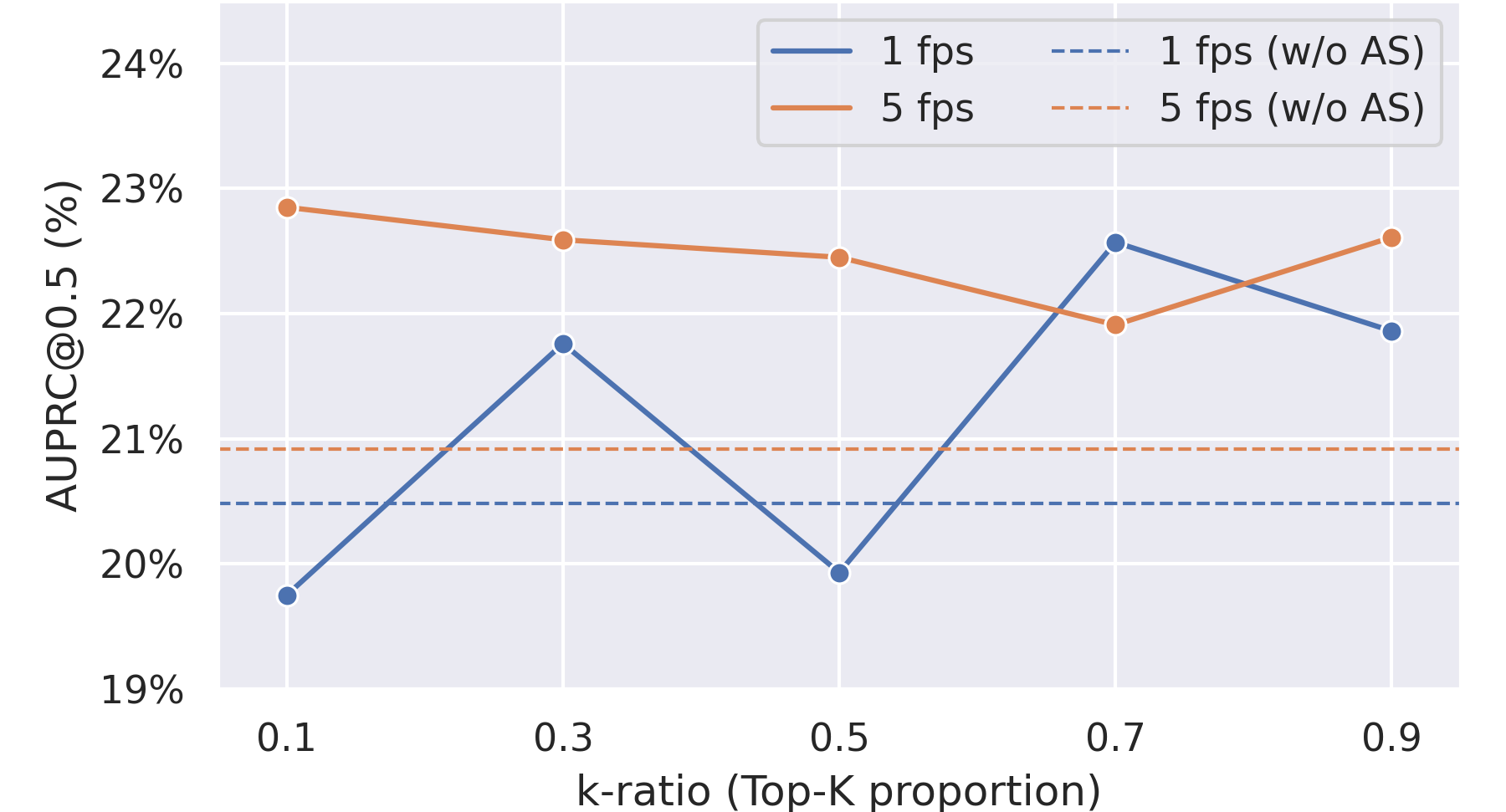} 
  \caption{Performance under different AS k-ratio at 1 fps and 5 fps (dashed = w/o AS).}
  \label{fig:k_ratio}
\end{figure}

\textbf{Redundancy reduction and regularization ablations.} Table~\ref{tab:as_reg_ablation} reports error-detection results when we vary the input video (feature) frame rate with and without the adaptive sampling module. All variants exclude view embeddings and bidirectional cross-fusion, instead using channel concatenation. The adaptive sampler consistently improves detection performance by removing redundant frames, improving temporal alignment, and enhancing sensitivity to temporal discrepancies. We also validate the effectiveness of the regularization terms (\(\mathcal{L}_{\mathrm{sel}}\) and \(\mathcal{L}_{\mathrm{vic}}\)) in aiding learning.
\begin{table}[htbp]
  \centering
  \small
  \setlength{\tabcolsep}{4pt}
  \begin{tabular}{lccccc}
  \toprule
  \multirow{2}{*}[-0.8ex]{\textbf{Variant}} & \multicolumn{4}{c}{\textbf{Input frame rate (fps)}} & \multirow{2}{*}[-0.8ex]{\textbf{Avg}} \\
  \cmidrule(lr){2-5}
   & 1 & 5 & 10 & 20 &  \\
  \midrule
  w/o AS              & 20.48 & 20.92 & 19.96 & 21.09 & 20.61 \\
  w AS   & 22.15 & 22.20 & \textbf{23.74} & 21.45 & 22.39 \\
  w AS, \(\mathcal{L}_{\mathrm{vic}}\), \(\mathcal{L}_{\mathrm{sel}}\)              & \textbf{23.51} & \textbf{22.60} & 22.63 & \textbf{22.65} & \textbf{22.85} \\
  \bottomrule
  \end{tabular}
  \caption{Results of the AS at different input frame rates and ablation of the regularization term. Metric is AUPRC@0.5.}
  \label{tab:as_reg_ablation}
  \end{table}
\\\textbf{Top-k analysis.} Fig.~\ref{fig:k_ratio} examines the impact of different top-k retention ratios (keeping the top k\% scoring frames). At lower frame rates, a larger top-k is preferable to avoid information loss; at higher frame rates, where redundancy and alignment mismatch are more pronounced, retaining a small subset of high-scoring frames suffices to improve performance.
\\\textbf{Frame-score visualization.} In Fig.~\ref{fig:performance}a, we visualize dynamic scores for Ego and Exo videos across diverse scenarios. Ego-frame saliency is notably more concentrated, which aligns with the human learning pattern of first closely observing the demonstration and then imitating a few salient, well-memorized key moments.
\subsubsection{SVE Analysis}
\begin{figure}[htbp]
  \centering
  \includegraphics[width=0.9\linewidth]{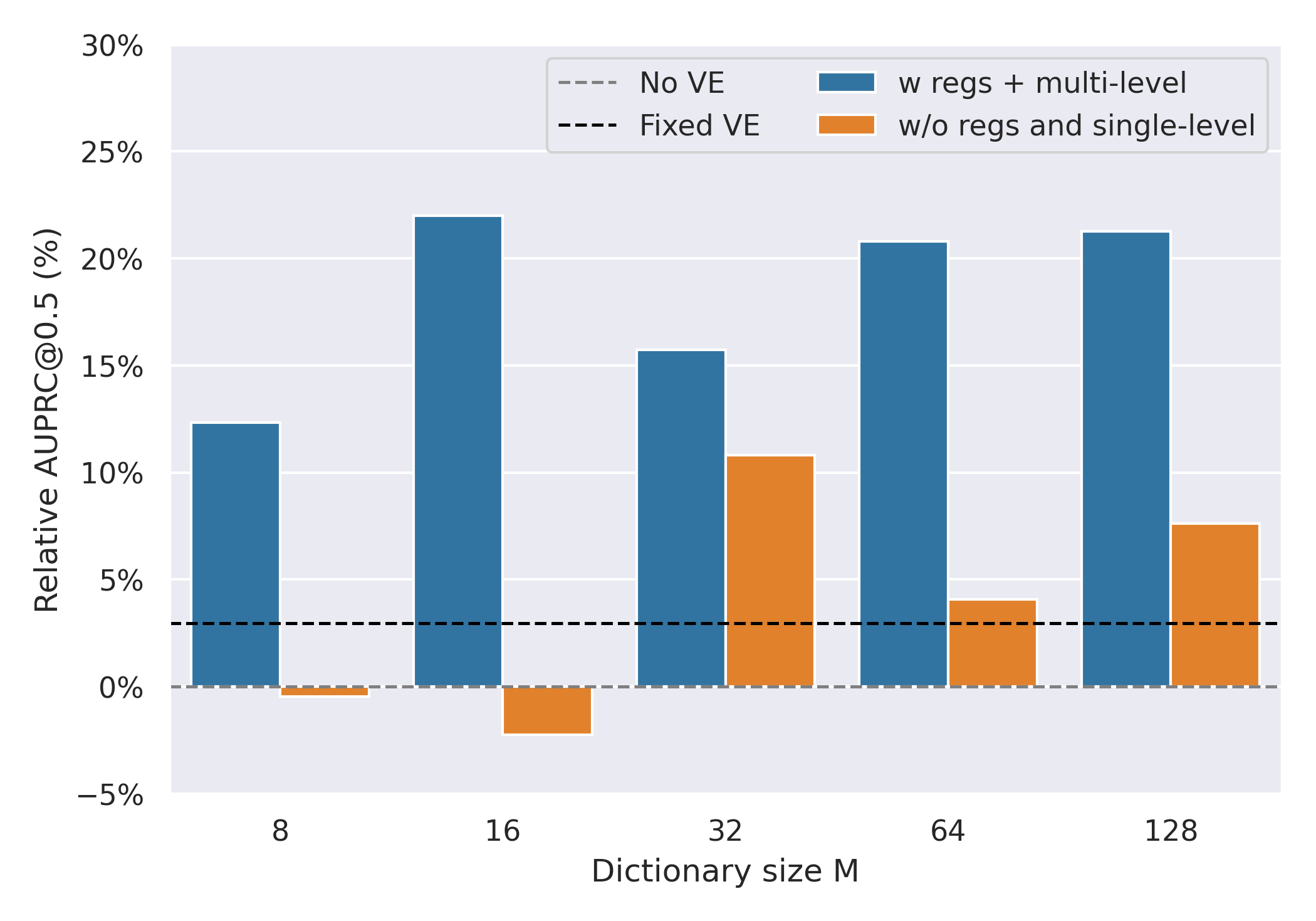} 
  \caption{Relative gain vs. dictionary size for scene-aware view embeddings on the EgoMe validation split. Dashed lines indicate baselines, gray one without view embeddings, black one with fixed learnable view embeddings.}
  \label{fig:SVE ablation}
\end{figure}

\begin{figure}[htbp]
  \centering
  \includegraphics[width=0.9\linewidth]{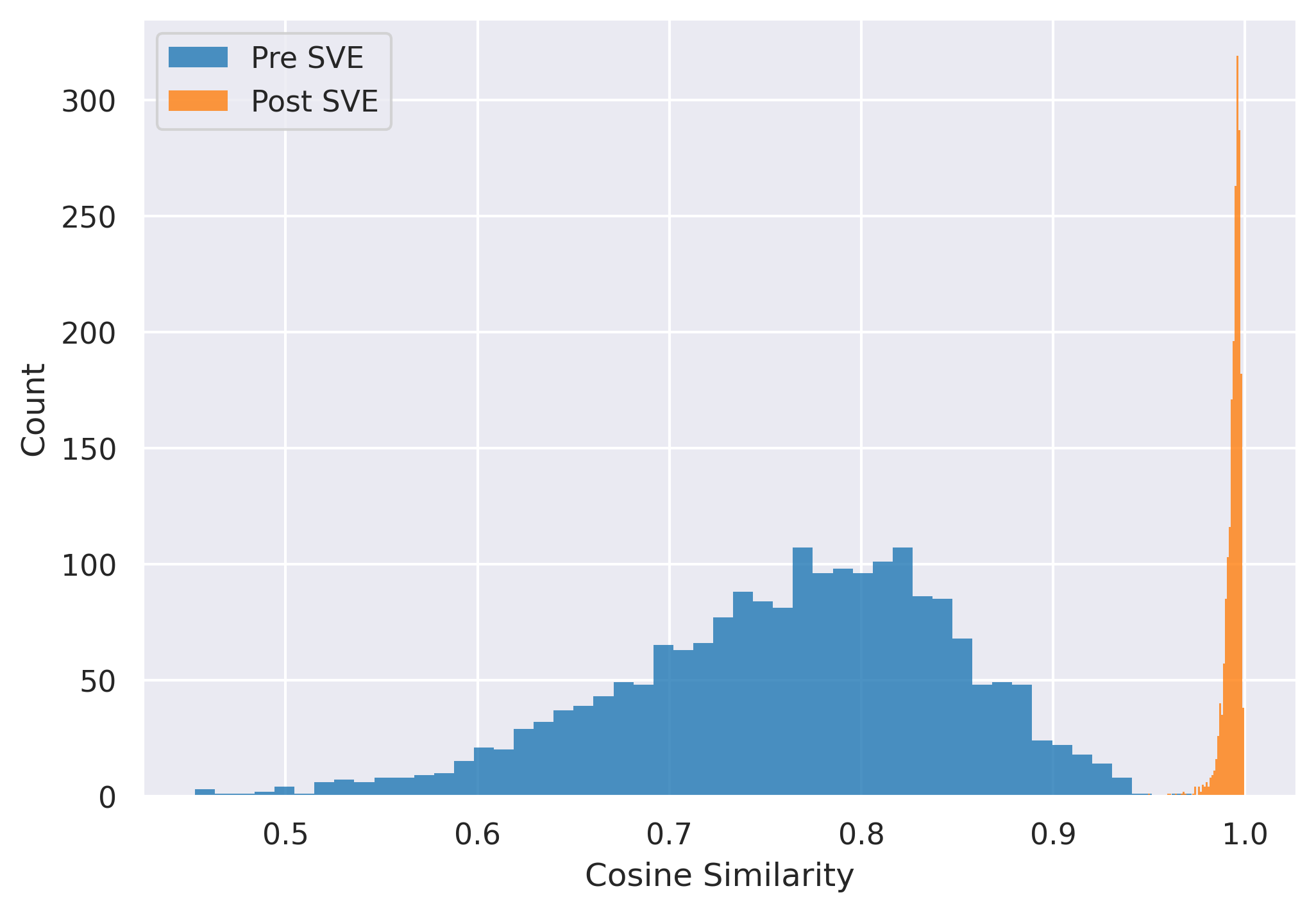} 
  \caption{Visualization of domain-discrepancy changes after SVE injection.}
  \label{fig:cosine_hist}
\end{figure}

\textbf{Comparison with fixed view embeddings}. We replace SVE with two learnable and test-time fixed tokens \(\mathbf{VE}^{exo}\) and \(\mathbf{VE}^{ego}\). As the black dashed line in Fig.~\ref{fig:SVE ablation} shows, gains are limited, whereas SVE delivers consistent improvements, implying fixed tokens cannot model cross-scene/view discrepancies effectively.
\\\textbf{Effect of scene-dictionary size}. Fig.~\ref{fig:SVE ablation} analyzes the influence of the dictionary size. We observe that moderately enlarging the dictionary better covers common view sub-factors, leading to more stable performance gains; when the dictionary is too small, limited expressiveness results in insufficient benefits.
\\\textbf{Ablation on regularization and multi-level injection}. We further ablate the roles of regularization and multi-level injection within the SVE module. Fig.~\ref{fig:SVE ablation} shows that, for each \(M\), adding regularizers such as selection-entropy/diversity mitigates over-sharpened attention and improves prototype coverage, while multi-level injection continuously modulates representations along the temporal hierarchy. In combination, these components enable SVE to consistently outperform fixed view embeddings across dictionary sizes and to deliver more uniform gains.
\\\textbf{Domain-discrepancy analysis}. We compute video-level global representations by uniformly pooling the Ego/Exo feature sequences over time before and after SVE injection. For each paired video, we then compute the cosine similarity between the two global representations and analyze its distribution (see Fig.~\ref{fig:cosine_hist}). With SVE, the similarity distribution shifts rightward and becomes more concentrated: the mean increases and the long tail is compressed, indicating that the cross-view domain gap is effectively mitigated.

\subsubsection{BiX Analysis}
\textbf{Comparison with alternative fusion schemes}. We also evaluate simpler fusion strategies—channel-wise concatenation and temporal sequence concatenation—and, for the attention module, compare global attention against deformable attention. The aggregated results are reported in Table~\ref{tab:fusion_ablation}.
\\\textbf{Ablation of bidirectional attention}. For the bidirectional cross-attention, we further decompose it into one-directional variants and report their performance separately (Table~\ref{tab:fusion_ablation}). We observe that the Exo\(\rightarrow\)Ego variant performs on par with the bidirectional setting, whereas Ego\(\rightarrow\)Exo is clearly weaker. This aligns with the task objective: our primary goal is error detection on the egocentric stream, so providing boundary and ordering cues from demonstration (Exo) to imitation (Ego) is more critical; the reverse direction is complementary but not decisive.

\begin{table}[!htbp]\small
  \centering
  \setlength{\tabcolsep}{4pt}
  \begin{tabular}{lccccc}
  \toprule
  \multirow{2}{*}[-0.8ex]{\textbf{Fusion Variant}} & \multicolumn{4}{c}{\textbf{AUPRC}} & \multirow{2}{*}[-0.8ex]{\textbf{tIoU}} \\
  \cmidrule(lr){2-5}
   & 0.3 & 0.5 & 0.7 & Mean & \\
  \midrule
  \multicolumn{6}{l}{\emph{Simple concatenation}} \\
  Concat (Channel)                       & 28.21 & 20.48 & 7.95 & 18.88 & 58.58 \\
  Concat (Time)                          & 28.91 & 21.05 & 7.85 & 19.27 & 58.15 \\
  \midrule
  \multicolumn{6}{l}{\emph{Attention Type}} \\
  BiX                   & \textbf{33.08} & 21.86 & 8.23 & \textbf{21.06} & 58.27 \\
  BiX-Deformable         & 32.99 & 21.05 & 8.56 & 20.87 & \textbf{58.92} \\
  \midrule
  \multicolumn{6}{l}{\emph{Single-direction ablations}} \\
  BiX (Exo$\rightarrow$Ego)              & 31.94 & \textbf{22.04} & 8.20 & 20.73 & 58.71 \\
  BiX (Ego$\rightarrow$Exo)              & 29.96 & 20.12 & \textbf{8.37} & 19.48 & 58.51 \\
  \bottomrule
  \end{tabular}
  \caption{Fusion variants on EgoMe validation split. \textbf{Concat (Channel)} concatenates Ego/Exo along the feature channel; \textbf{Concat (Time)} concatenates along the temporal axis. \textbf{BiX} denotes bidirectional cross-attention fusion; \textbf{BiX-Deformable} replaces cross-attention with deformable attention. The last block ablates the bidirectional mechanism into single-direction flows.}
  \label{tab:fusion_ablation}
\end{table}
\section{Conclusion}
We presented SAVA-X for Ego→Exo imitation error detection, addressing redundancy, cross-view domain gaps, and temporal misalignment through adaptive sampling, scene-adaptive view embeddings, and bidirectional cross-attention. On the EgoMe benchmark, SAVA-X consistently outperforms strong dense video captioning and temporal action localization baselines, and ablation studies show that each component targets a distinct bottleneck while yielding complementary gains when combined. Additional analyses on dictionary size, regularization, and fusion variants clarify the design trade-offs and failure modes of the framework. We hope that our unified protocol, baselines, and architecture will serve as a useful reference point for future work on cross-view imitation analysis and error detection in procedural tasks.

\section*{Acknowledgements}
This work was supported by the National Natural Science Foundation of China (No.~U23A20286 and No.~62301121), Sichuan Science and Technology Program  (No.~2026NSFSC1478) and Postdoctoral Fellowship Program (Grade B) of China Postdoctoral Science Foundation (No.~2025M783502 and No.~GZB20240120).
{
    \small
    \bibliographystyle{ieeenat_fullname}
    \bibliography{main}
}

\clearpage
\renewcommand{\thesection}{\Alph{section}}
\renewcommand{\thesubsection}{\thesection.\arabic{subsection}}

\renewcommand{\thefigure}{\thesection.\arabic{figure}}
\renewcommand{\thetable}{\thesection.\arabic{table}}
\renewcommand{\theequation}{\thesection.\arabic{equation}}
\setcounter{section}{0}
\setcounter{figure}{0}
\setcounter{table}{0}
\setcounter{equation}{0}
\maketitlesupplementary
\section{Implementation Details}
\subsection{Metric Computation}
\paragraph{AUPRC computation.}
For a fixed temporal IoU threshold, we collect all predicted segments and their confidence scores $\{s_i\}_{i=1}^N$ together with binary labels $\{y_i\}_{i=1}^N$ indicating whether each prediction is a true positive ($y_i=1$) or a false positive ($y_i=0$). Let $P$ denote the total number of positive ground-truth instances. Following the COCO protocol, we first construct a precision--recall curve from these predictions and then approximate the area under the curve (AUPRC) by averaging the interpolated precision values at 101 uniformly sampled recall points in $[0,1]$. This yields an interpolated average precision, which we report as AUPRC. In our experiments, we apply this procedure separately for different ``positive'' definitions (e.g., correct vs.\ error events) by treating each target category as positive and all others as negative.

\begin{algorithm}[t]
\caption{COCO-style AUPRC computation for a fixed tIoU}
\label{alg:auprc}
\begin{algorithmic}[1]
\REQUIRE Scores $s_i$, labels $y_i \in \{0,1\}$ for $i=1,\dots,N$; number of positives $P = \sum_i y_i$
\ENSURE AUPRC value $\mathrm{AP}$
\STATE Sort indices $\pi$ such that $s_{\pi_1} \ge s_{\pi_2} \ge \dots \ge s_{\pi_N}$
\STATE Initialize cumulative true/false positives: $\mathrm{TP}[k] = \sum_{i=1}^{k} \mathbb{1}[y_{\pi_i}=1]$, $\mathrm{FP}[k] = \sum_{i=1}^{k} \mathbb{1}[y_{\pi_i}=0]$
\FOR{$k = 1$ to $N$}
  \STATE $\mathrm{precision}[k] \leftarrow \frac{\mathrm{TP}[k]}{\max(1, \mathrm{TP}[k] + \mathrm{FP}[k])}$
  \STATE $\mathrm{recall}[k] \leftarrow \frac{\mathrm{TP}[k]}{\max(1, P)}$
\ENDFOR
\STATE For $k = N-1$ down to $1$:
\STATE \hspace{1.5em} $\mathrm{precision}[k] \leftarrow \max\big(\mathrm{precision}[k], \mathrm{precision}[k+1]\big)$
\STATE Initialize $\mathrm{AP} \leftarrow 0$
\FOR{$j = 0$ to $100$}
  \STATE $r_j \leftarrow j / 100$ \hfill \COMMENT{101 uniformly sampled recall points}
  \STATE Find the smallest index $k$ such that $\mathrm{recall}[k] \ge r_j$
  \IF{such $k$ exists}
    \STATE $p(r_j) \leftarrow \mathrm{precision}[k]$
  \ELSE
    \STATE $p(r_j) \leftarrow 0$
  \ENDIF
  \STATE $\mathrm{AP} \leftarrow \mathrm{AP} + p(r_j)$
\ENDFOR
\STATE $\mathrm{AP} \leftarrow \mathrm{AP} / 101$
\STATE \textbf{return} $\mathrm{AP}$
\end{algorithmic}
\end{algorithm}

\subsection{Base Encoder}
Our base encoder follows the multi-scale temporal convolutional encoder in Wang et al.~\cite{wang_endtoend_2021}. Given frame visual features $\mathbf{v}\in\mathbb{R}^{T\times d}$ from the backbone, we first reshape them to a 1D feature map of size $T\times d$ and feed them into a lightweight temporal pyramid. The first level uses a $1\times1$ Conv1D and GroupNorm to project the input feature dimension to the transformer hidden dimension. Subsequent levels apply $3\times3$ Conv1D with stride~2 and GroupNorm to progressively downsample the sequence, producing multi-scale features with decreasing temporal resolution and shared hidden dimension. For each level, we add a sine-based temporal positional encoding and pass the resulting feature maps and masks to the Deformable Transformer encoder. This design keeps the hidden sizes and number of feature levels consistent with \cite{wang_endtoend_2021}, so that performance gains mainly stem from our cross-view modules rather than changes in the base encoder. 
\subsection{Deformable DETR}
Our event detection head is a 1D multi-scale Deformable DETR that follows the design of Wang et al.~\cite{wang_endtoend_2021}. Given the multi-scale temporal features and masks from the base encoder, we first flatten all levels into a single sequence and add level-specific embeddings and temporal positional encodings. The encoder then applies $L_{\mathrm{enc}}$ layers of multi-scale deformable self-attention and feed-forward networks to produce a unified feature memory. The decoder takes a fixed set of learnable query embeddings and performs, at each of its $L_{\mathrm{dec}}$ layers, (i) self-attention among queries and (ii) multi-scale deformable cross-attention to the encoded memory using normalized reference points, followed by a feed-forward network. As in the original Deformable DETR, we use iterative reference-point refinement via a small MLP head attached to each decoder layer, and share the hidden dimension and attention configuration with \cite{wang_endtoend_2021}. This keeps the detection head identical to the PDVC Deformable DETR implementation, so that the performance gains in our experiments mainly arise from the proposed cross-view modules rather than changes in the underlying transformer architecture.

\begin{figure*}[t]
  \centering
  \includegraphics[width=0.85\linewidth]{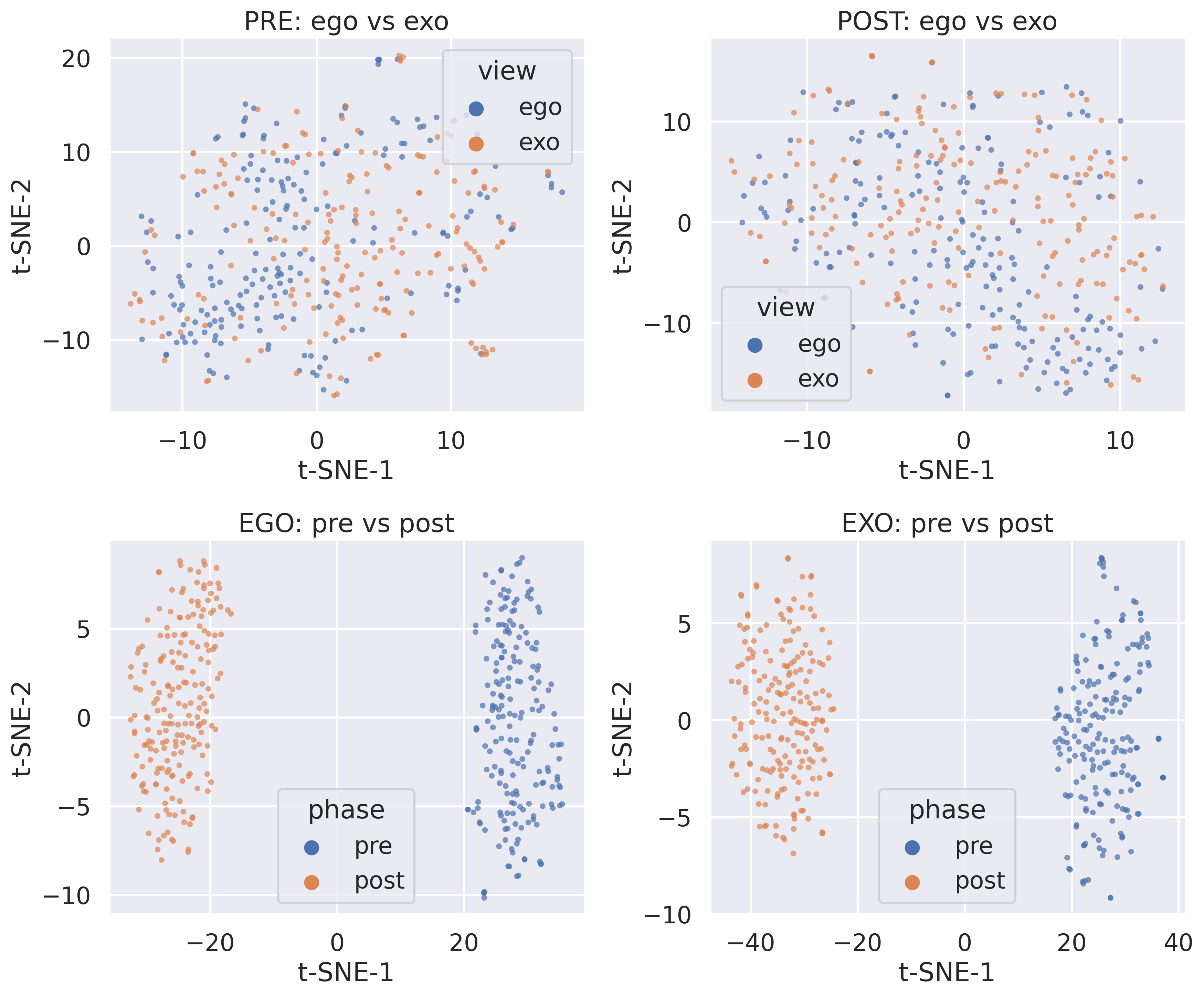} 
  \caption{t-SNE visualization of video-level features before and after SVE. \textbf{Top}: ego and exo features colored by view; SVE reduces the cross-view gap and yields more overlapped distributions. \textbf{Bottom}: ego (left) and exo (right) features colored by phase (pre vs.\ post), showing that SVE applies a structured, non-trivial transformation while avoiding representational collapse.}
  \label{fig:tsne}
\end{figure*}
\subsection{Task-specific heads}
\label{app:heads}
Except for the imitation-error prediction heads introduced in our work, all task heads follow Wang et al.~\cite{wang_endtoend_2021}.
Given the decoder features for each query, we apply a linear classification head to predict the event category and a regression MLP to predict the normalized start/end coordinates of the temporal segment.

Following PDVC, we additionally use a count head to estimate the number of events and a captioning head to generate a natural-language description for each detected segment, sharing the same hidden dimension and decoder outputs as in \cite{wang_endtoend_2021}.
On top of these standard heads, we attach two lightweight binary imitation-error heads:
(i) a fine-grained query-level error head that predicts whether each event segment corresponds to a correct or erroneous execution step, and
(ii) a global video-level head that aggregates query features to predict the overall imitation quality of the sequence.
These additions leave the original PDVC heads unchanged, ensuring that performance gains mainly stem from our cross-view alignment and error modeling rather than modifications to the baseline detection and captioning heads.

\subsection{Losses}
\label{app:loss}
\noindent\textbf{Dense video captioning loss.}
Following PDVC~\cite{wang_endtoend_2021}, we treat dense video captioning as a set prediction problem and supervise all decoder layers with a Hungarian-matching loss.
Let $\{\hat{\mathbf{s}}_i,\hat{\mathbf{c}}_i,\hat{\mathbf{y}}_i\}_{i=1}^N$ denote the predicted temporal segments (center--length parameterization), foreground scores, and caption word distributions for $N$ event queries in one decoder layer, and let $\{\mathbf{s}_j,\mathbf{y}_j\}_{j=1}^{N_\text{gt}}$ be the ground-truth segments and captions.
We first solve a bipartite matching between predictions and ground truths with cost
\begin{equation}
  \mathcal{C}_{ij}
  = \alpha_{\mathrm{giou}}\,\mathcal{L}_{\mathrm{giou}}(\hat{\mathbf{s}}_i,\mathbf{s}_j)
  + \alpha_{\mathrm{cls}}\,\mathcal{L}_{\mathrm{cls}}(\hat{\mathbf{c}}_i,\mathbb{1}[j\le N_\text{gt}])
\end{equation}
where $\mathcal{L}_{\mathrm{giou}}$ is the temporal generalized IoU loss and $\mathcal{L}_{\mathrm{cls}}$ is the focal classification loss between foreground/background.
Given the optimal assignment $\sigma$, the DVC loss for one decoder layer is
\begin{equation}
  \begin{aligned}
  \mathcal{L}_{\mathrm{DVC}}
  &= \beta_{\mathrm{giou}}\,\mathcal{L}_{\mathrm{giou}}
   + \beta_{\mathrm{cls}}\,\mathcal{L}_{\mathrm{cls}}
   + \beta_{\mathrm{ec}}\,\mathcal{L}_{\mathrm{ec}}
   + \beta_{\mathrm{cap}}\,\mathcal{L}_{\mathrm{cap}}\\
  \mathcal{L}_{\mathrm{giou}}
  &= \frac{1}{N_\text{gt}}\sum_{j=1}^{N_\text{gt}}
     \bigl(1 - \mathrm{GIoU}(\hat{\mathbf{s}}_{\sigma(j)},\mathbf{s}_j)\bigr)\\
  \mathcal{L}_{\mathrm{cls}}
  &= \frac{1}{N}\sum_{i=1}^{N}
     \mathcal{L}_{\mathrm{focal}}\big(\hat{\mathbf{c}}_i,\,y_i^{\mathrm{fg/bg}}\big)\\
  \mathcal{L}_{\mathrm{ec}}
  &= -\log p_{\mathrm{ec}}\big(N_\text{gt}\big)\\
  \mathcal{L}_{\mathrm{cap}}
  &= \frac{1}{N_\text{gt}}\sum_{j=1}^{N_\text{gt}}
     \frac{1}{T_j}\sum_{t=1}^{T_j}
     -\log p\big(w_{j,t}\mid w_{j,<t},\hat{\mathbf{z}}_{\sigma(j)}\big)
  \end{aligned}
\end{equation}
where $p_{\mathrm{ec}}$ is the event-counter distribution over event numbers, $\hat{\mathbf{z}}_{\sigma(j)}$ is the matched query feature, and $w_{j,t}$ is the $t$-th word of the $j$-th ground-truth caption of length $T_j$.
As in~\cite{wang_endtoend_2021}, we attach prediction heads to all decoder layers and sum the layer-wise losses.

\begin{table*}[t]\small
  \centering
  \begin{tabular*}{0.95\textwidth}{@{\extracolsep{\fill}} l*{5}{c}
    @{}@{\hspace{8pt}}!{\vrule width .8pt}@{\hspace{2pt}}@{}
    *{5}{c}}
  \toprule
  \multirow{2}{*}[-0.8ex]{\textbf{Method}}
   & \multicolumn{5}{c}{\textbf{AUPRC on Validation}}
   & \multicolumn{5}{c}{\textbf{AUPRC on Test}} \\
  \cmidrule(lr){2-6} \cmidrule(lr){7-11}
   & 0.3 & 0.5 & 0.7 & Mean & tIoU
   & 0.3 & 0.5 & 0.7 & Mean & tIoU \\
  \midrule
  \multicolumn{11}{@{\extracolsep{\fill}}l}{\emph{Dense Video Captioning (DVC) baselines}} \\
  PDVC~\cite{wang_endtoend_2021}
   & 68.41 & 46.21 & 12.72 & 42.45 & 58.58
   & \textbf{66.53} & 43.88 & 12.30 & 40.90 & 57.98 \\
  Exo2EgoDVC~\cite{ohkawa_exo2egodvc_2024}
   & 67.52 & 44.52 & 12.43 & 41.49 & 59.06
   & 65.37 & 42.27 & 11.40 & 39.68 & 58.15 \\
  \midrule
  \multicolumn{11}{@{\extracolsep{\fill}}l}{\emph{Temporal Action Localization (TAL) baselines}} \\
  ActionFormer~\cite{zhang_actionformer_2022}
   & 65.87 & 31.68 & 4.54 & 34.03 & 48.89
   & 63.25 & 29.20 & 4.17 & 32.20 & 48.25 \\
  TriDet~\cite{shi_tridet_2023}
   & 65.05 & 32.25 & 4.35 & 33.88 & 49.05
   & 62.45 & 30.29 & 4.30 & 32.35 & 49.02 \\
  \midrule
  \multicolumn{11}{@{\extracolsep{\fill}}l}{\emph{Only Egocentric Input}} \\
  PDVC~\cite{wang_endtoend_2021}
   & 64.77 & 42.11 & 10.94 & 39.27 & 57.63
   & 63.94 & 40.96 & 12.03 & 38.98 & 57.19 \\
  \midrule
  \multicolumn{11}{@{\extracolsep{\fill}}l}{\emph{Ours}} \\
  SAVA-X
   & \textbf{69.02} & \textbf{46.58} & \textbf{13.85} & \textbf{43.15} & \textbf{59.31}
   & 66.32 & \textbf{43.98} & \textbf{12.41} & \textbf{40.90} & \textbf{58.32} \\
  \bottomrule
  \end{tabular*}
  \caption{Comparison on EgoMe validation and test split. Left: results on \emph{validation set}. Right: results on the \emph{test set}. We report AUPRC for the correct class at multiple tIoU thresholds (0.3, 0.5, 0.7), their mean, and standalone temporal IoU (tIoU) for localization quality.}
  \label{tab:egome_all_vs_erroronly_correct}
\end{table*}

\noindent\textbf{Imitation-error classification losses.}
On top of the standard DVC loss, we introduce two lightweight error-prediction heads: a fine-grained query-level head and a global video-level head (Sec.~\ref{app:heads}).
For each matched query $j$, let $\hat{z}^{\mathrm{fine}}_{\sigma(j)}$ be the scalar logit from the fine-grained error head and $e^{\mathrm{fine}}_j\in\{0,1\}$ be the corresponding binary error label (1 for erroneous execution, 0 for correct).
We supervise this head with a binary cross-entropy (BCE) loss over the matched events:
\begin{equation}
  \begin{gathered}
  \mathcal{L}_{\mathrm{err}}^{\mathrm{fine}}
  = \frac{1}{N_\text{gt}}\sum_{j=1}^{N_\text{gt}}
    \Big[-e^{\mathrm{fine}}_j\log\sigma\!\big(\hat{z}^{\mathrm{fine}}_{\sigma(j)}\big)\\
         -(1-e^{\mathrm{fine}}_j)\log\big(1-\sigma\!\big(\hat{z}^{\mathrm{fine}}_{\sigma(j)}\big)\big)\Big]
  \end{gathered}
\end{equation}
where $\sigma(\cdot)$ is the sigmoid function.
In addition, the global head aggregates all valid query features of a video into a single logit $\hat{z}^{\mathrm{overall}}$ that predicts the overall imitation quality, with binary label $e^{\mathrm{overall}}\in\{0,1\}$ (error-free vs.\ containing errors).
We apply another BCE loss:
\begin{equation}
  \begin{gathered}
  \mathcal{L}_{\mathrm{err}}^{\mathrm{overall}}
  = -e^{\mathrm{overall}}\log\sigma\!\big(\hat{z}^{\mathrm{overall}}\big)\\
    -(1-e^{\mathrm{overall}})\log\big(1-\sigma\!\big(\hat{z}^{\mathrm{overall}}\big)\big)
    \end{gathered}
\end{equation}

\noindent\textbf{Adaptive sampling regularization.}
We further regularize the adaptive sampler (AS) to avoid collapsed selections and redundant representations.
Denote by $\{s_{x,t}\}_{t=1}^{T_x}$ and $\{s_{y,t}\}_{t=1}^{T_y}$ the normalized selection probabilities over Ego/Exo tokens.
We add a selection-entropy regularizer~\cite{pereyra2017regularizing} that encourages coverage instead of concentrating mass on a few positions:
\begin{equation}
  \begin{gathered}
  \mathcal{L}_{\mathrm{sel}}
  = \frac{1}{\log T_x}\sum_{t=1}^{T_x}s_{x,t}\,\log\!\big(s_{x,t}+\varepsilon\big)\\
  + \frac{1}{\log T_y}\sum_{t=1}^{T_y}s_{y,t}\,\log\!\big(s_{y,t}+\varepsilon\big)
  \end{gathered}
  \label{eq:sel_entropy_app}
\end{equation}
where $\varepsilon>0$ is a small constant for numerical stability.
Let $u\in\{\mathrm{exo},\mathrm{ego}\}$ index the view and $\hat{\mathbf{Z}}^{u}\in\mathbb{R}^{K_u\times d}$ be the matrix of gated active tokens (after selection and gating), with $K_u$ tokens and feature dimension $d$.
We further attach VICReg-style~\cite{bardes2022vicreg} variance and covariance penalties to suppress collapse and dimensional collinearity:
\begin{equation}
  \begin{gathered}
  \boldsymbol{\mu}^{u}
  = \frac{1}{K_u}\sum_{i=1}^{K_u}\hat{\mathbf{Z}}^{u}_{i},\quad
  \hat{\mathbf{Z}}^{u}_{\mathrm{c}}
  = \hat{\mathbf{Z}}^{u}-\mathbf{1}\,{\boldsymbol{\mu}^{u}}^{\!\top}\\
  \mathcal{L}^{u}_{\mathrm{var}}
  = \frac{1}{d}\sum_{j=1}^{d}
    \Big[\max\!\big(0,\ \gamma - \sqrt{\mathrm{Var}(\hat{\mathbf{Z}}^{u}_{\mathrm{c},j})+\varepsilon}\big)\Big]^2\\
  \mathbf{C}^{u}
  = \frac{1}{K_u-1}\,\hat{\mathbf{Z}}_{\mathrm{c}}^{u\top}\hat{\mathbf{Z}}^{u}_{\mathrm{c}},\quad
  \mathcal{L}^{u}_{\mathrm{cov}}
  = \frac{1}{d}\sum_{\substack{i=1\\ i\neq j}}^{d}\sum_{j=1}^{d}\big(\mathbf{C}^{u}_{ij}\big)^2\\
  \mathcal{L}_{\mathrm{vic}}
  = \mathcal{L}^{\mathrm{exo}}_{\mathrm{var}}+\mathcal{L}^{\mathrm{ego}}_{\mathrm{var}}
  + \mathcal{L}^{\mathrm{exo}}_{\mathrm{cov}}+\mathcal{L}^{\mathrm{ego}}_{\mathrm{cov}}
  \end{gathered}
  \label{eq:vicreg_app}
\end{equation}
where $\gamma>0$ is the variance lower bound and $\varepsilon>0$ again ensures numerical stability.

\section{Results}

\subsection{Correct Class Results}
Table~\ref{tab:egome_all_vs_erroronly_correct} summarizes the AUPRC performance for the \emph{correct} (non-error) class on EgoMe validation and test splits.

Among DVC-style baselines, PDVC~\cite{wang_endtoend_2021} remains strong, while Exo2EgoDVC~\cite{ohkawa_exo2egodvc_2024} does not consistently improve over PDVC despite explicitly transferring exocentric knowledge. TAL-only models (ActionFormer~\cite{zhang_actionformer_2022}, TriDet~\cite{shi_tridet_2023}) perform clearly worse on AUPRC and tIoU, showing that off-the-shelf localization architectures are not sufficient for our fine-grained imitation setting.

Using only egocentric input further degrades PDVC, highlighting the benefit of multi-view information even when evaluating the correct class.

Our SAVA-X achieves the best validation performance across all tIoU thresholds and mean AUPRC, and attains comparable or better test performance than PDVC, with particularly noticeable gains at high tIoU (e.g., $+1.1$ AUPRC@0.7 on validation). These results indicate that SAVA-X not only improves error detection, but also preserves or slightly enhances recognition and localization of correct executions, rather than trading off one class for the other.

\subsection{TSNE}
To better understand how SVE reshapes the feature space, we visualize pre- and post-SVE video-level features using t-SNE (Fig.~\ref{fig:tsne}).

In the top row, we color points by view (ego vs.\ exo).
Before SVE (top-left), ego and exo features form two partially separated clouds with a clear domain shift.

After SVE (top-right), the two distributions become much more interleaved, indicating that SVE effectively reduces the cross-view gap and encourages a more view-invariant embedding.
In the bottom row, we fix the view and color points by phase (pre vs.\ post).

For both ego (bottom-left) and exo (bottom-right), pre- and post-SVE features form two well-separated clusters along the first t-SNE dimension, showing that SVE applies a non-trivial, structured transformation to the representations rather than a small perturbation.

Together, these plots suggest that SVE consistently aligns ego and exo distributions while preserving meaningful intra-view variability and avoiding collapse.

\end{document}

%% file: preamble.tex
\usepackage{pifont}
\newcommand{\cmark}{\ding{51}} 
\usepackage{multirow}
\usepackage{graphicx}
\usepackage[margin=1in]{geometry}
\usepackage{algorithm}
\usepackage{algorithmic}
\usepackage{booktabs}
\usepackage{array} 
\usepackage[dvipsnames,svgnames]{xcolor} 

\graphicspath{{figures/}{imgs/}}  

\newcommand{\gain}[1]{\textbf{\textcolor{SeaGreen}{#1}}} 

